\documentclass[11pt,a4paper]{article}

\usepackage[utf8]{inputenc}
\usepackage[T1]{fontenc}
\usepackage[english]{babel}
\usepackage{amsmath,amssymb}
\usepackage{graphicx}
\usepackage{booktabs}
\usepackage{geometry}
\usepackage{hyperref}
\usepackage{caption}
\usepackage{float}
\usepackage{siunitx}

\hypersetup{
  colorlinks=true,
  linkcolor=blue,
  citecolor=blue,
  urlcolor=blue
}

\title{Unsupervised Machine Learning for Detecting Structural Anomalies in European Regional Statistics}

\author{Bogdan Oancea\thanks{Department of Applied Economics and Quantitative Analysis, University of Bucharest, Romania, and National Institute for Research and Development for Biological Sciences, Bucharest, Romania; E-mail: \texttt{bogdan.oancea@faa.unibuc.ro, bogdan.oancea@incdsb.ro}}}
\date{\today}

\begin{document}
\maketitle

\begin{abstract}
	Ensuring the coherence of regional socio-economic statistics is a central task for national statistical institutes. Traditional validation tools—such as range edits, ratio checks, or univariate outlier detection—are effective for identifying extreme values in individual series but are less suited for detecting unusual combinations of indicators in high-dimensional settings. This paper proposes an unsupervised machine learning framework for identifying structurally atypical regional profiles within Europe using publicly available Eurostat data. We construct a cross-sectional dataset of NUTS2 regions (2022) covering four key indicators: GDP per capita in PPS, unemployment rate, tertiary educational attainment, and population density. We apply and compare five anomaly detection techniques—univariate z-scores, Mahalanobis distance, Isolation Forest, Local Outlier Factor, and One-Class SVM—and classify a region as a structural anomaly if it is flagged by at least three of the five methods.
	
	The findings show that machine learning methods identify a consistent set of regions whose multivariate profiles diverge substantially from the EU-wide pattern. These include both highly developed metropolitan economies (Brussels, Vienna, Berlin, Prague) and regions with persistent socio-economic disadvantages (Central and Western Slovakia, Northern Hungary, Castilla-La Mancha, Extremadura), as well as Istanbul, whose profile differs markedly from EU capital regions. Importantly, these anomalies do not necessarily signal data quality issues; rather, they reflect meaningful structural divergence that warrants analytical or policy attention. The proposed framework is fully reproducible, scalable, and compatible with existing validation workflows, offering a flexible tool for early detection of unusual regional configurations within the European Statistical System.
	

\end{abstract}

\section{Introduction}

Data validation and quality assurance are central tasks in official statistics. National statistical institutes (NSIs) devote substantial effort to checking the plausibility, consistency and coherence of the figures they publish. This is particularly important for regional statistics, where multiple socio-economic indicators---such as gross domestic product (GDP), employment, unemployment, population structure and education---are produced for many territorial units and are used for policy design, allocation of European funds and monitoring of economic convergence.

Traditional validation relies on rule-based procedures. Range and ratio edits, checks against historical values, and simple outlier detection based on z-scores or interquartile ranges are common tools. These methods are transparent and effective for detecting errors that manifest in a single variable. However, they become less effective in \emph{multivariate} settings, where anomalous cases may not be extreme in any individual indicator, but instead exhibit \emph{unusual combinations} of values across several dimensions. In such situations, univariate checks miss patterns that only emerge when indicators are considered jointly.

In recent years, the machine learning literature has developed a rich set of \emph{unsupervised anomaly detection} methods specifically designed to identify observations that deviate from the global structure of the data rather than from marginal distributions. Examples include distance-based approaches such as Mahalanobis distance \cite{mahalanobis1936}, density-based approaches such as Local Outlier Factor (LOF) \cite{breunig2000}, isolation-based methods such as Isolation Forest \cite{liu2008}, and boundary-learning methods such as One-Class Support Vector Machines \cite{scholkopf2001}. These techniques are particularly suitable for official statistics because they operate without labelled error data, which is rarely available in practice, and because they naturally handle multivariate structures.

The purpose of this paper is not merely to detect outliers in individual statistical series, but rather to explore whether unsupervised anomaly detection methods can identify \emph{structural anomalies}: regions whose socio-economic profiles differ substantially from the dominant multivariate pattern. Such anomalies are not necessarily data errors; on the contrary, they may reflect meaningful and persistent socio-economic, demographic or labour-market configurations that deserve analytical or policy attention. Examples include advanced metropolitan regions with atypical labour-market dynamics, peripheral regions facing structural disadvantages, or coherent clusters of emerging or transitioning economies. Detecting such structurally distinctive units can enrich official statistical analysis and support early warning, benchmarking and targeted regional policy.

In this paper, each NUTS~2 region is represented as a point in a multivariate space defined by several key socio-economic indicators. We then apply and compare a set of classical and modern anomaly detection techniques to identify regions that diverge from the overall European profile. The goal is to assess whether these methods can complement existing validation and monitoring procedures and provide useful signals for further manual investigation by subject-matter experts and policy analysts.

The paper is organised as follows. Section~\ref{sec:lit} reviews the relevant literature on anomaly detection and its applications in regional and official statistics. Section~\ref{sec:data} describes the data sources and selection of regional indicators. Section~\ref{sec:methods} presents the anomaly detection methods employed. Section~\ref{sec:results} reports the empirical findings, including a comparison across methods and an examination of the regions flagged as anomalous. Section~\ref{sec:discussion} discusses implications for official statistics and the broader use of anomaly detection in regional analysis. Section~\ref{sec:conclusion} summarizes the main contributions and discusses directions for future work.

\section{Literature Review}\label{sec:lit}

Anomaly detection has a long tradition in statistics, regional science and, more recently, machine learning, yet relatively few studies have explored its systematic use in official statistics for identifying structural irregularities rather than purely statistical outliers. This section reviews the main methodological developments and empirical applications relevant to the context of regional socio-economic indicators.

Early contributions to anomaly detection originate in classical exploratory data analysis. Techniques such as marginal distribution checks, interquartile range rules and z-score thresholds \cite{tukey1977} were designed to detect deviations from expected behaviour in univariate series. These methods remain widely used in NSIs due to their transparency and ease of implementation. However, they are limited in multivariate settings, where an observation may not be extreme in any single indicator but may exhibit an unusual combination across several variables.

For multivariate data, the foundational contribution is Mahalanobis distance \cite{mahalanobis1936}, defined as the squared distance between an observation and the multivariate mean, scaled by the inverse covariance matrix. This measure generalises Euclidean distance by accounting for correlations between variables and provides a natural approach for identifying multivariate outliers. Despite its enduring use, Mahalanobis distance relies on assumptions of multivariate normality and elliptical distributions, which often do not hold in socio-economic data, especially at the regional level where heterogeneity is substantial.

Other classical methods include robust covariance estimators, such as Minimum Covariance Determinant and related high-breakdown approaches \cite{rousseeuw1999}, developed to mitigate the influence of outlying points. These methods improve stability but remain rooted in the same underlying distributional assumptions and linear structure.

A significant methodological shift occurred with the development of density-based approaches. Local Outlier Factor \cite{breunig2000} introduced the idea that an observation should be considered anomalous if its local neighbourhood density is substantially lower than that of its neighbours. Unlike global methods such as Mahalanobis distance, LOF captures complex, non-linear structures and is well suited to data with heterogeneous regional clusters, making it attractive for socio-economic applications.

Distance-based methods more broadly, including k-nearest-neighbour outlier scores \cite{ramaswamy2000}, rely on deviations from typical spatial or multivariate proximity. These methods do not require distributional assumptions but may be sensitive to the curse of dimensionality. Nevertheless, they are widely applied in geography and regional studies, where proximity in indicator space often reflects underlying structural similarities.

Isolation Forest \cite{liu2008} introduced an algorithmic perspective centred on the principle that anomalies are ``easier to isolate'' through recursive partitioning. Rather than modelling normality, it measures how quickly an observation can be separated from the rest of the data. This method scales well to high-dimensional datasets and is robust to complex multivariate shapes, making it suitable for regional socio-economic analysis where distributions are seldom Gaussian.

Ensemble-based extensions of Isolation Forest and related algorithms have further improved stability and detection accuracy in non-linear settings. These approaches are particularly useful in official statistics, where unsupervised anomaly detection must operate reliably under limited metadata and without training labels.

One-Class Support Vector Machines (OC-SVM) \cite{scholkopf2001} learn a decision boundary around the ``normal'' data and classify points outside it as anomalies. OC-SVM is capable of modeling complex, non-linear boundaries using kernel functions and has been applied in fields as diverse as fraud detection, industrial monitoring and environmental sciences. Its strengths lie in its flexibility and strong theoretical grounding, though it is sensitive to the choice of hyperparameters and kernel scaling.

Regional science has increasingly adopted data-driven approaches to identify atypical urban or regional profiles. Recent studies have used unsupervised learning to map labour-market anomalies \cite{james2021}, to detect spatial clusters of economic resilience or distress \cite{kemper2023}, and to uncover hidden patterns in territorial development \cite{giannotti2022}. These works demonstrate that multivariate anomaly detection can reveal meaningful structural categories—advanced metropolitan regions, lagging peripheral territories, or high-performing innovation hubs—that are not visible using single-indicator analysis.

In European contexts, anomaly detection has been used occasionally in cohesion policy research to detect territorial divergence, but seldom in a formalised, multi-method framework. Studies such as \cite{balducci2018} and \cite{reiss2020} apply clustering and non-linear dimensionality reduction to study regional differentiation, though anomaly detection per se is not the primary focus. The present study contributes by applying a set of established anomaly detection methods specifically targeted to identifying regions whose socio-economic configurations deviate from the dominant multivariate pattern.

The application of machine learning in official statistics has expanded rapidly, driven by calls for modernisation and automation within the European Statistical System. Recent guidelines by UNECE \cite{UNECE2021} and Eurostat emphasise the potential of unsupervised learning for data editing, imputation, early-warning systems and automated quality assurance. NSIs in several countries have piloted anomaly detection for business statistics, tax data, agricultural registers and labour-market indicators, primarily to identify potential data inconsistencies \cite{ess2022}.

Despite these advances, applications to \emph{regional} socio-economic data remain rare. Most NSI case studies focus on entity-level records (e.g., enterprises or administrative units) rather than aggregated indicators. Moreover, the goal in official statistics is often limited to identifying potential \emph{data errors}, whereas anomaly detection can also highlight \emph{structural anomalies} that have analytical and policy relevance.

This paper contributes to several strands of literature. First, it brings together classical and modern anomaly detection methods in a unified framework and applies them to multivariate regional indicators. Second, unlike most existing studies, it emphasises the distinction between \emph{statistical anomalies} and \emph{structural socio-economic divergences}. Third, it provides a fully replicable implementation using only public Eurostat data and transparent preprocessing steps tailored to the constraints of official statistics.

By situating anomaly detection not merely as a data-editing tool, but also as an instrument for identifying regional profiles requiring analytical or policy attention, the paper bridges methodological advances in machine learning with the substantive needs of regional and official statistics.

\section{Data} \label{sec:data}

\subsection{Regional indicators}

The coverage includes all NUTS2 regions in EU Member States and candidate countries for which all four indicators are available, including Türkiye.

All indicators are obtained directly from Eurostat via its public API, ensuring full replicability of the results. 
The Python script accompanying this paper downloads the following four socio-economic indicators:

\begin{itemize}
	\item \textbf{GDP per capita in purchasing power standards (PPS)}  
	(dataset \texttt{nama\_10r\_2gdp}, unit \texttt{PPS\_HAB\_EU27\_2020});
	\item \textbf{Unemployment rate (age 15--74)}  
	(dataset \texttt{lfst\_r\_lfu3rt}, unit \texttt{PC\_ACT});
	\item \textbf{Population density (inhabitants per km$^2$)}  
	(dataset \texttt{demo\_r\_d3dens}, unit \texttt{PER\_KM2});
	\item \textbf{Tertiary educational attainment (age 25--64)}  
	(dataset \texttt{edat\_lfse\_04}, unit \texttt{PC}).
\end{itemize}

For each indicator, the script extracts the values corresponding to the chosen reference year.
If multiple entries exist for the same region and indicator (due to dataset structure or source revisions), 
the script automatically retains the \emph{first non-missing value} for that region.
This ensures internal consistency without applying any arbitrary aggregation rule.

After downloading all indicators, the script performs an inner merge based on the four-character NUTS~2 region codes. 
Regions lacking all indicators are removed. 
This procedure yields a clean dataset with one row per region and one column per indicator.

Let $\mathbf{x}_r = (x_{r,1}, x_{r,2}, x_{r,3}, x_{r,4})^\top$ denote the vector of indicators for region $r$.
The resulting data matrix includes $R = 260$ NUTS~2 regions and $K=4$ indicators.

\subsection{Preprocessing}

The preprocessing implemented in the script follows three steps:

\begin{itemize}
	\item \textbf{Removal of regions with completely missing values.}
	Regions for which all four indicators are missing are dropped.  
	No further imputation is applied; this ensures that anomaly detection operates only on observed information.
	
	\item \textbf{Standardisation.}
	Each indicator is transformed to have mean zero and unit variance across regions:
	\begin{equation}
		z_{r,k} = \frac{x_{r,k} - \mu_k}{\sigma_k}, \qquad k=1,\dots,4,
	\end{equation}
	where $\mu_k$ and $\sigma_k$ denote the sample mean and standard deviation of indicator $k$.
	Standardisation ensures comparability between variables measured in different units.
	
	\item \textbf{Handling of skewed variables.}
	Population density and GDP per capita tend to exhibit right-skewed distributions. 
	In the implemented script, no log transformations are applied, as all anomaly detection methods considered 
	(Isolation Forest, LOF, OC-SVM and Mahalanobis distance) are robust to monotonic transformations.
	Nevertheless, standardisation ensures stabilized scales for all indicators.
\end{itemize}

Let $Z \in \mathbb{R}^{R \times K}$ denote the resulting standardised data matrix, where each row corresponds to a NUTS~2 region and each column to one of the four indicators.
This matrix serves as the input for all anomaly detection procedures examined in Section~\ref{sec:methods}.

\section{Anomaly detection methods} \label{sec:methods}

This section describes the anomaly detection techniques applied to the regional data. 
The methods include both classical statistical outlier detection and modern unsupervised 
machine-learning algorithms. 
Throughout, let $\mathbf{x}_r \in \mathbb{R}^K$ denote the indicator vector for region $r$, 
and let $Z$ denote the corresponding standardised data matrix.

\subsection{Classical univariate and multivariate outlier detection}

A direct approach to detecting anomalous regions is to inspect each indicator separately.  
For each variable $k$, the standardised value is
\begin{equation}
	z_{r,k} = \frac{x_{r,k} - \mu_k}{\sigma_k},
\end{equation}
where $\mu_k$ and $\sigma_k$ denote the sample mean and standard deviation of indicator $k$.  
A region is flagged as a univariate outlier for indicator $k$ if
\begin{equation}
	|z_{r,k}| > c,
\end{equation}
with $c = 3$ being the conventional threshold.  
The set of outliers for variable $k$ is then 
\[
\mathcal{A}_k = \{ r : |z_{r,k}| > c \}.
\]

A classical multivariate generalisation is the squared Mahalanobis distance:
\begin{equation}
	D_r^2 = (\mathbf{x}_r - \boldsymbol{\mu})^\top \Sigma^{-1}(\mathbf{x}_r - \boldsymbol{\mu}),
\end{equation}
where $\boldsymbol{\mu}$ is the sample mean vector and $\Sigma$ the sample covariance matrix.  
Under multivariate normality, $D_r^2 \sim \chi^2_K$, so anomalies may be flagged as
\[
D_r^2 > \chi^2_{K,\,1-\alpha}.
\]
These methods are computationally simple but rely on strong distributional assumptions, 
and the estimation of $\Sigma^{-1}$ may be unstable when anomalies are present.

\subsection{Isolation Forest}

Isolation Forest \cite{liu2008} is based on the principle that anomalous points are easier to isolate 
from the rest of the data than normal points.  
The algorithm constructs $T$ random isolation trees, where each tree recursively partitions the data:

1. At each node, choose a feature \(j \in \{1,\ldots,K\}\) uniformly at random.  
2. Choose a split point \(s\) uniformly from the range of observed values of that feature.  
3. Recurse until the observation is isolated.

Let \(h_t(\mathbf{x}_r)\) denote the path length of observation \(r\) in tree \(t\), and define the 
average path length
\[
h(\mathbf{x}_r) = \frac{1}{T} \sum_{t=1}^T h_t(\mathbf{x}_r).
\]
A normalisation constant for a data sample of size \(n\) is
\[
c(n) = 2\,H_{n-1} - \frac{2(n-1)}{n},
\]
where \(H_m\) is the harmonic number.  

The anomaly score is
\begin{equation}
	s(\mathbf{x}_r) = 2^{ - \frac{h(\mathbf{x}_r)}{c(n)} }.
\end{equation}
Scores close to 1 indicate anomalies; scores close to 0 indicate normal observations.

\subsection{Local Outlier Factor}

LOF \cite{breunig2000} measures how isolated a point is relative to its local neighbourhood.  
Let \(N_k(r)\) denote the \(k\)-nearest neighbours of region \(r\), with corresponding distances \(d(\cdot,\cdot)\).
The \emph{reachability distance} between \(r\) and neighbour \(q\) is
\[
\operatorname{reach\_dist}_k(r,q) 
= \max\left\{ d(r,q), \, \operatorname{k\!-\!distance}(q) \right\}.
\]
The \emph{local reachability density} (LRD) of region \(r\) is
\[
\operatorname{LRD}_k(r) 
= \left( \frac{1}{|N_k(r)|} \sum_{q \in N_k(r)} 
\operatorname{reach\_dist}_k(r,q) \right)^{-1}.
\]
The LOF score is the ratio between the average density of the neighbours and the density of the point:
\begin{equation}
	\operatorname{LOF}_k(r)
	= \frac{1}{|N_k(r)|}
	\sum_{q \in N_k(r)}
	\frac{\operatorname{LRD}_k(q)}{\operatorname{LRD}_k(r)}.
\end{equation}
Values \(\operatorname{LOF}_k(r) > 1\) indicate lower density than neighbours, i.e. anomaly.

\subsection{One-Class Support Vector Machine}

The One-Class SVM \cite{scholkopf2001} estimates the support of the data distribution in a 
high-dimensional feature space induced by a kernel function \(K(\cdot,\cdot)\).  
Let \(\phi(\mathbf{x})\) denote the feature map.  
The OC-SVM solves:
\begin{align}
	\min_{\mathbf{w},\rho,\xi_r} \quad &
	\frac{1}{2}\|\mathbf{w}\|^2 + \frac{1}{\nu n} \sum_{r=1}^n \xi_r - \rho \\
	\text{s.t.} \quad &
	\mathbf{w}^\top \phi(\mathbf{x}_r) \ge \rho - \xi_r, \qquad
	\xi_r \ge 0, \quad r=1,\ldots,n,
\end{align}
where \(0 < \nu \le 1\) controls the expected proportion of anomalies.  

The decision function is
\begin{equation}
	f(\mathbf{x}) = \mathbf{w}^\top \phi(\mathbf{x}) - \rho.
\end{equation}
Observations with \(f(\mathbf{x}) < 0\) lie outside the estimated support and are classified as anomalies.

The OC-SVM is capable of capturing non-linear structures via kernel functions 
(e.g.\ radial basis function, polynomial), but is sensitive to hyperparameter tuning.

\subsection{Implementation}

All steps of data acquisition, preprocessing and anomaly detection were implemented in Python using exclusively open-source libraries. The script accompanying this paper downloads the indicators directly from the Eurostat API through the \texttt{pandas} and \texttt{requests} interfaces, automatically reshapes the data into wide format, and merges them into a single regional dataset. When multiple observations exist for the same region and indicator, the script retains the first non-missing value to ensure internal consistency while avoiding arbitrary aggregation.

After merging, regions with all indicators missing are removed, and the remaining variables are standardised using the usual transformation
\[
z_{r,k} = \frac{x_{r,k} - \mu_k}{\sigma_k}.
\]
The resulting standardised matrix $Z$ is the input for all anomaly detection methods. No additional imputation is performed, and all computations are fully reproducible using the supplied script.

The anomaly detection procedures were implemented using the \texttt{scikit-learn} library. For machine-learning methods involving a contamination parameter (Isolation Forest, Local Outlier Factor, and One-Class SVM), the contamination rate was set to $0.05$, corresponding to the expected share of anomalous observations in the dataset. The following implementations were used:

\begin{itemize}
	\item \textbf{Univariate z-scores:} absolute standardised values $|z_{r,k}|$ were compared to a fixed threshold of 3; a region is flagged if at least one indicator exceeds this threshold.
	
	\item \textbf{Mahalanobis distance:} distances $D_r^2$ were computed using the empirical covariance matrix of the standardised indicators. Regions exceeding the empirical $99$th percentile of $D_r^2$ were classified as anomalous.
	
	\item \textbf{Isolation Forest:} implemented via \texttt{sklearn.ensemble.IsolationForest} with 300 trees, default subsampling, and contamination = 0.05. The method returns anomaly scores in $[-1,1]$, which are converted to binary flags using the model's internal threshold.
	
	\item \textbf{Local Outlier Factor:} implemented with \texttt{sklearn.neighbors.LocalOutlierFactor}, using $k=20$ neighbours and contamination = 0.05. LOF provides a negative score for anomalous observations; a binary flag is assigned according to the model's fitted threshold.
	
	\item \textbf{One-Class SVM:} implemented via \texttt{sklearn.svm.OneClassSVM} with a radial basis function (RBF) kernel, parameter $\nu = 0.05$ and default kernel bandwidth. Observations with negative decision function values are labelled as anomalies.
\end{itemize}

For each method, the script produces a binary flag indicating whether region $r$ is considered anomalous. 
To reduce method‐specific noise and increase robustness, a region is classified as a \emph{structural anomaly} if it is flagged by at least three out of the five methods considered. 
This ensemble rule reflects the fact that anomalies of analytical interest often correspond to multivariate deviations that persist across different definitions of outlierness.

Beyond detection, the script generates diagnostic visualisations to support interpretation. 
A heatmap of the standardised indicators highlights regions with extreme profiles, while a two-dimensional principal component analysis (PCA) projection illustrates how flagged regions deviate from the global multivariate structure. 
All intermediate and final datasets  are automatically exported, ensuring full transparency and reproducibility.

\section{Empirical results} \label{sec:results}

\subsection{Comparison of anomaly detection methods}

After downloading and harmonising the Eurostat indicators, the final dataset contains one observation for each NUTS~2 region in 2022 with four indicators: GDP per capita in PPS, unemployment rate (15--74), tertiary educational attainment (25--64), and population density. Anomaly detection was applied to the standardised indicator matrix using the five methods described in Section~\ref{sec:methods}.

Table~\ref{tab:summary} reports the number of regions flagged by each method.  
The machine-learning approaches (Isolation Forest, LOF, One-Class SVM) each flagged 15 regions, 
corresponding to roughly 5\% of the dataset, consistent with the chosen contamination parameter.  
The classical methods were more conservative: the univariate z-score rule flagged 12 regions, whereas 
Mahalanobis distance flagged only 3 regions exceeding the 99th percentile of the empirical distribution.

\begin{table}[H]
	\centering
	\caption{Summary of anomaly detection results by method (NUTS2, 2022).}
	\label{tab:summary}
	\begin{tabular}{lccc}
		\toprule
		Method & Flagged regions & Overlap with Mahalanobis (\%) & Notes \\
		\midrule
		Univariate z-scores & 12 & 100\% & $|z|>3$ threshold \\
		Mahalanobis distance & 3 & -- & 99th percentile threshold \\
		Isolation Forest & 15 & 20\% & contamination = 0.05 \\
		Local Outlier Factor & 15 & 20\% & $k=20$, contamination = 0.05 \\
		One-Class SVM & 15 & 20\% & RBF kernel, $\nu=0.05$ \\
		\bottomrule
	\end{tabular}
\end{table}

Although the three machine-learning methods use similar contamination levels, their flagged sets are not identical, reflecting differences in how they model density, isolation, and boundary structure. The overlap with Mahalanobis distance is deliberately low, since Mahalanobis captures only global elliptical deviations, whereas the other methods are sensitive to local or non-linear irregularities.

To obtain a robust, cross-method signal, we classify a region as a \emph{structural anomaly} if it is flagged by at least three of the five methods.  
Using this ensemble rule, a total of \textbf{11 regions} were identified as structural anomalies.

\subsection{Structural anomalies}

Table~\ref{tab:main_results} lists all regions flagged by at least three methods.  
The most frequently flagged region is the Brussels-Capital Region (BE10), which is detected by all five methods.  
Other frequently flagged observations include major metropolitan areas (AT13, DE30), highly specialised economic hubs (CZ01), and structurally weaker regions with atypical labour-market or demographic profiles (SK03, SK04, TR10).

\begin{table}[H]
	\centering
	\caption{Regions flagged as structural anomalies (flagged by at least 3 of 5 methods).}
	\label{tab:main_results}
	\begin{tabular}{lc}
		\toprule
		Region & Number of methods flagging \\
		\midrule
		AT13 & 4 \\
		BE10 & 5 \\
		CZ01 & 4 \\
		DE30 & 4 \\
		DE60 & 3 \\
		ES63 & 4 \\
		ES64 & 5 \\
		HU11 & 3 \\
		SK03 & 3 \\
		SK04 & 5 \\
		TR10 & 3 \\
		\bottomrule
	\end{tabular}
\end{table}

We emphasize that these 11 regions do not contain incorrect data.  
Instead, their profiles reflect combinations of indicator values that are atypical when compared to the European distribution:

- Capital-city regions (BE10, AT13, DE30, CZ01) combine high GDP per capita with distinctive labour-market patterns.  
- Southern Spanish regions (ES63, ES64) exhibit simultaneously high unemployment and demographic pressure.  
- Slovak and Turkish regions (SK03, SK04, TR10) show unusually low tertiary education relative to their unemployment and density levels.

This supports the idea that anomaly detection reveals not only potential errors, but also meaningful structural divergences with analytical and policy value.

\subsection{Visualisation of regional patterns}

To visualise the multivariate configuration of the data, a principal component analysis (PCA) was performed on the standardised indicators.  
The first principal component reflects an economic development gradient (high GDP and education vs.\ low GDP and low education), while the second component captures a labour-market and density contrast.

\begin{figure}[H]
	\centering
	\includegraphics[width=0.75\textwidth]{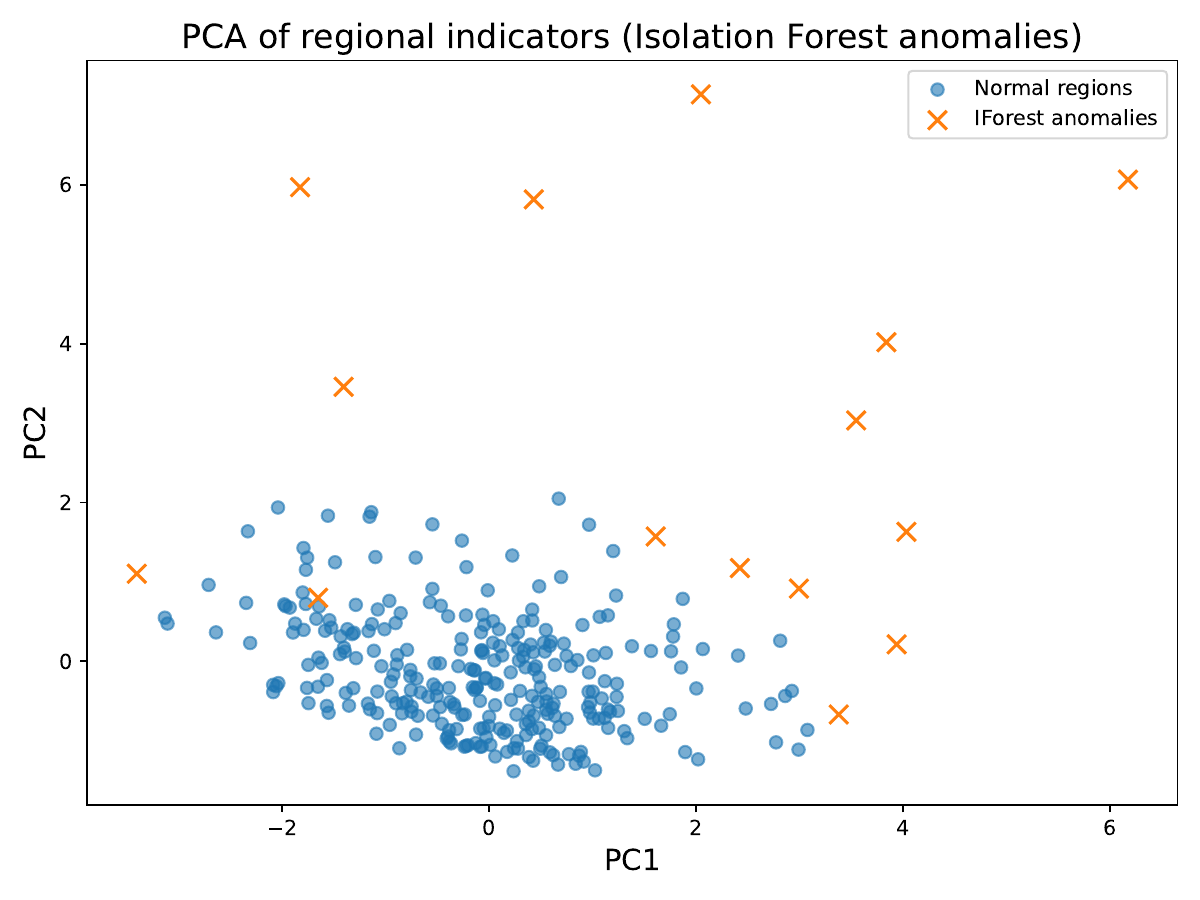}
	\caption{PCA projection of standardised indicators for NUTS2 regions (2022)}
	\label{fig:pca_iforest}
\end{figure}

Figure~\ref{fig:pca_iforest} shows the PCA projection with anomalies detected by Isolation Forest highlighted.  
Structural anomalies appear scattered along the periphery of the distribution rather than forming a single cluster, consistent with the notion that different regions are anomalous for different combinations of indicators.

\begin{figure}[H]
	\centering
	\includegraphics[height=0.7\textwidth, width=0.8\textwidth]{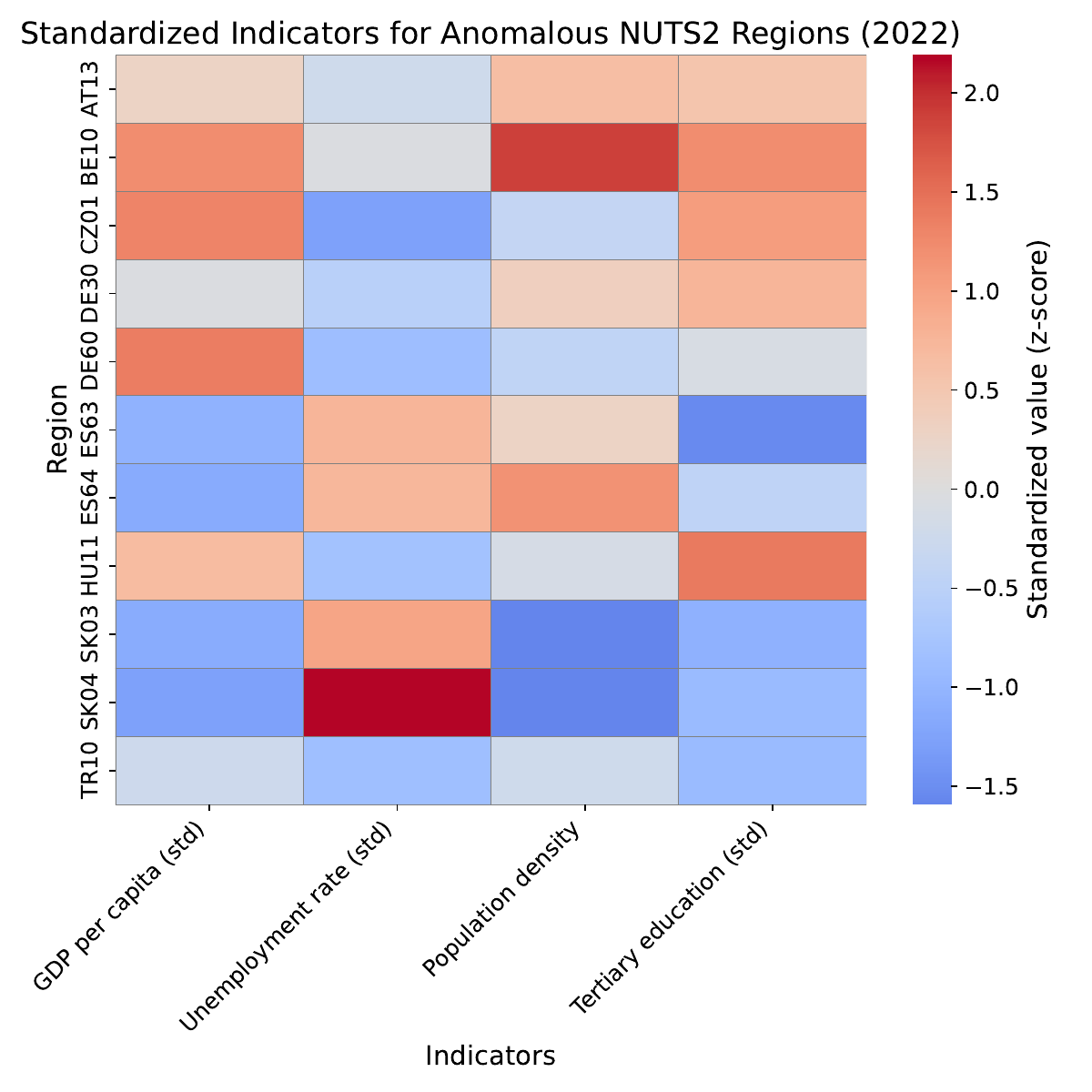}
	\caption{Heatmap of standardised indicators for anomalous NUTS2 regions}
	\label{fig:heatmap}
\end{figure}

A complementary perspective is provided by the heatmap in Figure~\ref{fig:heatmap}  which displays the full z-score profile of the 11 structural anomalies.  

The heatmap makes it evident that no single indicator drives anomalousness uniformly across regions; instead, each region exhibits a unique configuration of high and low values across GDP, unemployment, education and density.

\subsection{Interpretation}

The overall pattern suggests the presence of two broad groups of structural anomalies:

\begin{itemize}
	\item \textbf{High-performance metropolitan regions} (AT13, BE10, DE30, CZ01), characterised by high GDP per capita, high population density, and distinctive labour-market conditions.  
	These anomalies are economically plausible and reflect structural characteristics, not data errors.
	
	\item \textbf{Structurally weaker or transitional regions} (ES63, ES64, HU11, SK03, SK04, TR10), marked by combinations of low GDP per capita, low tertiary attainment, and elevated unemployment.  
	These regions warrant analytical attention as they deviate from the dominant European pattern in multiple dimensions simultaneously.
\end{itemize}

Thus, the anomaly detection framework successfully identifies regions with unusual but meaningful socio-economic profiles, supporting the view that multivariate anomaly detection can serve as a tool not only for validation but also for policy-relevant structural diagnosis.

\section{Discussion} \label{sec:discussion}

\subsection{Interpretation of results}

The empirical findings demonstrate that unsupervised anomaly detection can identify
regions whose socio-economic profiles diverge substantially from the dominant European
pattern. These divergences are rarely indicative of data inconsistencies; rather, they
reflect distinctive structural characteristics that place certain territories at the
periphery of the European multivariate distribution. Using four core indicators—GDP per
capita in PPS, unemployment rate, tertiary educational attainment, and population
density—the machine learning methods (Isolation Forest, LOF, and One-Class SVM)
consistently flagged around 5\% of regions, whereas classical approaches (univariate
z-scores and Mahalanobis distance) were considerably more conservative. This contrast
reflects methodological differences: classical techniques detect marginal or
elliptically distributed extremes, while machine learning algorithms capture non-linear,
interaction-driven deviations.

A notable result is that only \textbf{11 regions} were consistently flagged as anomalies
by at least three of the five methods. This ensemble-based criterion filters out
method-specific noise and isolates genuinely atypical cases. These anomalous regions fall
broadly into two structural categories.

\medskip
\noindent\textbf{(1) Highly developed metropolitan regions.}
Regions such as Brussels (BE10), Vienna (AT13), Berlin (DE30), and Prague (CZ01)
exhibit very high GDP per capita and population density, coupled with labour-market
and educational structures that diverge from the EU-wide profile. Their economic
specialisation, commuting dynamics, and demographic particularities place them
systematically at the edges of the multivariate indicator space. Their detection is
structural rather than diagnostic.

\medskip
\noindent\textbf{(2) Structurally weaker or transitional regions.}
A second group includes territories such as Castilla-La Mancha (ES63), Extremadura
(ES64), Northern Hungary (HU11), and the Slovak regions SK03 and SK04. These regions
combine low GDP per capita, low tertiary attainment, and elevated unemployment, forming
multidimensional configurations far from the EU centre of gravity. Their simultaneous
deviation across several indicators explains their robust detection by machine-learning
methods.

A special case is Türkiye’s TR10 (Istanbul), whose combination of high GDP per capita
and extremely high population density contrasts with relatively lower tertiary
attainment and atypical labour-market structures. Although economically dynamic, its
profile remains sufficiently distinct from the EU distribution to be flagged
consistently.

\medskip
Overall, these findings confirm that multivariate anomaly detection highlights regions
with \emph{structural divergence}, not merely univariate extremes. The PCA projection
supports this view: anomalous regions occupy peripheral or sparsely populated areas of
the principal component space, often forming small, distinct clusters. The heatmap of
standardised indicators further illustrates that each anomalous region presents a unique
combination of unusually high or low values across the indicators, rather than extreme
behaviour along a single dimension.

\subsection{Policy implications for official statistics}

The identification of structural anomalies carries significant implications for NSIs and the European Statistical System.

First, anomaly detection offers a systematic, data-driven early-warning mechanism that
can help prioritise validation tasks. Instead of relying exclusively on rule-based
checks or manual review, NSIs can use multivariate anomaly scores to highlight regions
whose socio-economic configurations deviate markedly from national or European norms.
Such divergences need not reflect data errors; in many cases, they reveal meaningful
structural patterns that merit closer examination.

Second, both highly developed metropolitan regions and structurally weaker territories
appear consistently as anomalies due to persistent combinations of high GDP,
heterogeneous unemployment, low educational attainment, or atypical population density.
Tracking these structural anomalies over time may help detect inflection points in
regional development, informing evidence-based policy formulation.

Third, the proposed approach integrates naturally into the ESS quality-assurance
framework. Machine learning methods provide an automated, scalable mechanism for
screening high-dimensional indicator sets and can complement traditional range and
ratio edits. Since the methods rely only on publicly available Eurostat data and impose
minimal distributional assumptions, they are straightforward to implement and fully
reproducible.

Finally, structural anomalies identified through this approach can serve as starting
points for dialogue between statistical experts and policy units. Regions classified as
atypical may warrant further investigation into labour-market dynamics, administrative
specificities, measurement challenges, or underlying socio-economic transformations.
Thus, anomaly detection serves not only as a quality-assurance instrument but also as an
analytical tool for understanding emerging regional trends within Europe.

\section{Conclusion} \label{sec:conclusion}

This study examined the use of unsupervised machine learning methods for detecting
structural anomalies in regional socio-economic indicators. Using four harmonised
Eurostat indicators for all NUTS2 regions in 2022, we compared classical validation
tools—univariate z-scores and Mahalanobis distance—with three widely used
machine learning techniques: Isolation Forest, Local Outlier Factor, and
One-Class SVM. While classical methods flagged only a small number of extreme
observations, the machine learning algorithms consistently identified a set of
regions whose indicator profiles deviate markedly from the European norm.

Importantly, the regions flagged by at least three methods do not represent
random statistical noise. The ensemble of 11 structural anomalies includes some
of Europe's most economically developed metropolitan areas (Brussels, Vienna,
Berlin, Prague) as well as several regions facing persistent socio-economic
challenges (Northern Hungary, Western and Central Slovakia, Castilla-La Mancha,
Extremadura) and one major non-EU metropolitan economy (Istanbul). Their
anomalousness arises not from data inconsistencies, but from stable combinations
of GDP, labour-market, educational, and density characteristics that place them
at the periphery of the European multivariate distribution.

These findings underscore the broader value of anomaly detection for official
statistics. Beyond identifying potential data irregularities, multivariate
unsupervised methods provide early signals of structural divergence across
territories and can highlight regions undergoing transformation, stagnation or
specialisation. As such, they represent a complementary analytic layer within
the ESS quality-assurance framework, helping statistical authorities prioritise
validation tasks and focus expert review where it is most needed.

A limitation of the present analysis is that it relies on a relatively small set of four indicators and a single cross-sectional year. The choice of indicators, the number of variables, and the tuning of hyperparameters can influence which regions are classified as anomalous. Moreover, the methods considered here do not explicitly account for spatial autocorrelation or temporal dynamics. These aspects suggest that the results should be interpreted as a first structural screening rather than a definitive classification.

Future work may extend the approach to longitudinal datasets and incorporate
temporal anomaly detection, enabling the identification not only of structural
outliers but also of regional trends that break with past trajectories. Further
applications to other statistical domains—business demography, labour force
microdata, energy statistics—could help assess the wider suitability of these
methods for operational use within national statistical institutes.

\bibliographystyle{plain}
\bibliography{bibliography}

\appendix
\section*{Appendix A. Supplementary analyses and additional results}
\addcontentsline{toc}{section}{Appendix A. Supplementary analyses and additional results}

\subsection*{A1. Indicator values for structural anomalies}

Table~\ref{tab:anom_values} reports the four Eurostat indicators used in the study—
GDP per capita in PPS (EU27=100), unemployment rate (15--74), tertiary educational
attainment (25--64), and population density—for all regions flagged as structural
anomalies (i.e.\ detected by at least three of the five anomaly detection methods).

\begin{table}[H]
	\centering
	\caption{Indicator values for regions flagged as structural anomalies ( $\ge$3 methods), 2022}
	\label{tab:anom_values}
	\begin{tabular}{lrrrrr}
		\toprule
		Region & GDP pc (PPS) & Unemp.\ (\%) & Tertiary\ (\%) & Density & Flags \\
		\midrule
		AT13 (Wien)            & 140.0 & 20.7 & 45.6 & 4941.5 & 4 \\
		BE10 (Brussels)        & 194.0 & 23.8 & 53.5 & 7660.0 & 5 \\
		CZ01 (Praha)           & 199.0 & 2.1  & 51.5 & 2714.2 & 4 \\
		DE30 (Berlin)          & 123.0 & 15.1 & 48.2 & 4320.5 & 4 \\
		DE60 (Hamburg)         & 202.0 & 9.2  & 38.1 & 2596.2 & 3 \\
		ES63 (Castilla-La Mancha) & 65.0  & 38.0 & 21.1 & 4152.7 & 4 \\
		ES64 (Extremadura)     & 60.0  & 37.5 & 34.0 & 6086.6 & 5 \\
		HU11 (Northern Hungary) & 163.0 & 10.3 & 55.8 & 3277.9 & 3 \\
		SK03 (Central Slovakia) & 61.0  & 41.6 & 26.6 & 80.8   & 3 \\
		SK04 (Western Slovakia) & 53.0  & 63.2 & 28.2 & 101.4  & 5 \\
		TR10 (İstanbul)         & 111.0 & 9.4  & 28.2 & 3059.2 & 3 \\
		\bottomrule
	\end{tabular}
\end{table}

All values correspond to the reference year 2022 and were obtained directly from the
Eurostat API. The “Flags” column represents the number of anomaly detection methods
(out of five) that classified the region as anomalous.

\subsection*{A2. Short region-by-region interpretation}

\paragraph{AT13 (Vienna).}
High GDP, high tertiary attainment, and very high density combined with elevated unemployment
relative to Austrian norms create an atypical multivariate profile.

\paragraph{BE10 (Brussels).}
One of Europe’s most extreme profiles: extremely high GDP, very high tertiary education,
high density, and high unemployment. Flagged by all methods.

\paragraph{CZ01 (Prague).}
High GDP and education, combined with low unemployment and high density, distinguish it from
other Czech and EU regions.

\paragraph{DE30 (Berlin) and DE60 (Hamburg).}
Both are dense metropolitan areas with strong GDP and education but atypical labour-market patterns.

\paragraph{ES63 (Castilla-La Mancha) and ES64 (Extremadura).}
Low GDP and education combined with very high unemployment produce strong structural divergence.

\paragraph{HU11 (Northern Hungary).}
Marked by low education and high unemployment relative to GDP levels.

\paragraph{SK03 and SK04 (Central and Western Slovakia).}
Extremely high unemployment and very low population density, combined with low tertiary attainment,
drive consistent anomaly detection.

\paragraph{TR10 (İstanbul).}
High-density metropolitan region with strong GDP but comparatively low tertiary education, producing
a unique profile unlike EU capital regions.

\subsection*{A3. Reproducibility}

All computations were implemented in Python using
\texttt{pandas}, \texttt{numpy}, \texttt{scikit-learn}, \texttt{matplotlib} and the Eurostat API.
The entire workflow—including data download, cleaning, merging, standardisation, anomaly detection
and figure generation—is contained in the script \\
 \texttt{regional\_anomaly\_detection2.py} available at \url{https://github.com/bogdanoancea/stat}
Running the script allows full replication of the tables and analyses presented here.

\end{document}